\documentclass[conference]{llncs}
\usepackage{cite}
\usepackage{multicol}
\usepackage{comment}
\usepackage{enumerate}
\usepackage{amsmath,amssymb,amsfonts}
\usepackage{algorithmic}
\usepackage{graphicx}
\usepackage{textcomp}
\usepackage{amssymb}
\usepackage{amsmath}
\usepackage{comment}
\usepackage{booktabs} % for better-looking horizontal lines
\usepackage{caption} % for better caption formatting
\usepackage[utf8]{inputenc}
\usepackage[english]{babel}
 \usepackage{mhchem} 
 \usepackage{multirow}
 \usepackage{multicol}
\usepackage{comment}
\usepackage{balance}
\usepackage{pgfplots}
\usepackage{tikz}
\usepackage{graphicx}
\usepackage{caption}
\usepackage{subcaption}
\usepackage{xcolor}
\def\BibTeX{{\rm B\kern-.05em{\sc i\kern-.025em b}\kern-.08em
    T\kern-.1667em\lower.7ex\hbox{E}\kern-.125emX}}

\begin{document}

\title{Enhancing Eye Disease Diagnosis with Deep Learning and Synthetic Data Augmentation}
\author{Saideep Kilaru \inst{1} \and Kothamasu Jayachandra \inst{1} \and Tanishka Yagneshwar \inst{1} \and Suchi Kumari$^*$ \inst{1}}
%\authorrunning{Kilaru and Kumari et al.}
\institute{Department of Computer Science Engineering, Shiv Nadar Institute of Eminence, Delhi-NCR, India. \email{deeps2657@gmail.com},
\email{jayachandra.kothamasu18@gmail.com},
\email{tanishka.yag@gmail.com},
\email{suchi.singh24@gmail.com}}

\authorrunning{S. Kumari et al. (Corresponding Author)}

\maketitle
%\institute {
%Department of Computer Science Engineering, Bennett University, Greater Noida, Uttar Pradesh, India, \email{samya.muhuri@bennett.edu.in}
%\and
%\maketitle

\begin{abstract}
In recent years, the focus is on improving the diagnosis of diabetic retinopathy (DR) using machine learning and deep learning technologies. Researchers have explored various approaches, including the use of high-definition medical imaging, AI-driven algorithms such as convolutional neural networks (CNNs) and generative adversarial networks (GANs). Among all the available tools, CNNs have emerged as a preferred tool due to their superior classification accuracy and efficiency. Although the accuracy of CNNs is comparatively better but it can be improved by introducing some hybrid models by combining various machine learning and deep learning models. Therefore, in this paper, an ensemble learning technique is proposed for early detection and management of DR with higher accuracy. The proposed model is tested on the APTOS dataset and it is showing supremacy on the validation accuracy ($99\%)$ in comparison to the previous models. Hence, the model can be helpful for early detection and treatment of the DR, thereby enhancing the overall quality of care for affected individuals.
\end{abstract}
\textbf{keyword: Diatetic retinopathy, ensemble learning, convolution neural network, eye disease diagnosis, APTOS }\\

\section{Introduction}
Among all the sense organs, eyes are considered as a crucial tool for human observation and perception to real worlds. The exposure of eyes with external space leads to the increasing chance of contact with external pathogens. Some viruses are entering to human body through the eyes, causing eye disease and affecting the visual functions. Sometimes these diseases or some eye injury cause visual impairment and blindness. It presents significant challenges in the lives of the affected people to focus on their needs. Researchers from various domains are providing various approaches to enhance the quality of life for the visually impaired people. Timely and accurate diagnosis of eye conditions can lead to prompt treatment and improved prognosis. Therefore, in the current era, researchers analyzed artificial intelligence driven technologies in the diagnosis of eye diseases, the development of visual aids, assistive devices, etc.

Due to the advancement in the modern medical imaging technology, the high-definition medical images play an important role in the disease diagnosis. Fundus camera is used to capture such images and can also be used to capture the complete structure of the retinal images such as vessels, optic cup, disc, macula, etc. Analyzing structural changes aids in identifying various diseases including Diabetic Retinopathy (DR) \cite{fong2003diabetic}, Diabetic Macular Edema (DME), Glaucoma, and others. Among all the diseases, DR is the most common disease, caused due to the increasing rate of diabetic patients in the recent past. According to WHO \cite{WHO}, approx.  422 million people worldwide have diabetes, the majority living in low-and middle-income countries. Early screening for DR is critical in preventing irreversible blindness, and this can be accomplished through manual examination or the application of intelligent tools on fundus images. Intelligent screening not only streamlines the process and reduces manpower requirements but also facilitates large-scale DR screening [4]. Wang \textit{et al.} \cite{ wang2023artificial} provides an overview of the recent research works in artificial intelligence-driven eye disease diagnosis and assistive technologies for the visually impaired human beings. It outlines potential future directions where artificial intelligence could further benefit the visually impaired. 

In recent years, medical image processing based on deep learning has become an important research direction, among which DR is an important research domain. The accuracy of the diagnostic model heavily depends on the input images as well as the labeling done by some professional doctors. The high-quality images can be generated automatically using an unsupervised adversarial generation model named as Generative Adversarial Network (GAN) \cite{yi2019generative}. It can generate any number of high-quality images through the random distribution of noise. However, GAN, is considered as inefficient and uncontrollable for complex data. A controlled variant of GAN \cite{mirza2014conditional} is proposed by applying some constraints to the generator and discriminator for pre-training of the data. Liang \textit{et al.} \cite{liang2022end} proposed an end to end CGAN by applying constraint on the loss function, namely class feature loss and retinal detail loss. 

While GANs offer capabilities for image generation and data augmentation, deep learning based model, particularly convolutional neural network (CNN) is considered as a preferred tool for predicting diabetic retinopathy due to its superior classification accuracy, interpretability, efficiency, and many more. \cite{ pratt2016convolutional} proposed a CNN approach for diagnosing  severity of DR from digital fundus image with validation accuracy of $75\%$ on Kaggle dataset. Lam et al. \cite{lam2018automated} used CNN and applied transfer learning on pretrained GoogLeNet and AlexNet models to automatically detect DR, and multiple fundus image datasets, and achieved an accuracy rate of $75.5\%$. A deep residual network (ResNet) model \cite{he2016deep} is also widely used for the diagnosis of DR. Wan \textit{et al.}  \cite{wan2018deep} achieved accuracy rate of 95.68\% by employing transfer learning architectures, including AlexNet, VGGNet, GoogleNet, and ResNet. They utilized publicly available data from the Kaggle platform for their study. Tufail et al. \cite{tufail2021diagnosis} used 3D convolutional neural network (3D-CNN) architectures for binary and multiclass (5 classes) classification of DR. Macsik \textit{et al.} \cite{macsik2022local} proposed local binary convolutional neural network (LBCNN) deterministic filter generation which can do diagnosis less learnable parameters and also with less memory usage in comparison with CNN. 

Multiple hybrid models are also proposed considering different deep learning (DL) models, and then the processed result is feed some classifiers for classification. Ali \textit{et al.} \cite{ali2023hybrid} used two deep learning models, Resnet50 and Inceptionv3, for preprocessing the images and then feed them into the CNN for classification. Nahiduzzaman \textit{et al.} \cite{nahiduzzaman2021hybrid} utilized Ben Graham’s technique and contrast limited adaptive histogram equalization (CLAHE) for preprocessing DR images. They then introduced a novel hybrid CNN-singular value decomposition (SVD) model to reduce input features for classifiers, followed by employing an Extreme Learning Machine (ELM) algorithm for classification to minimize training time.  Bilal \textit{et al.} \cite{bilal2024improved} employed a multi-stage strategy encompassing data preprocessing, feature extraction using a hybrid CNN-SVD model, and classification utilizing Improved Support Vector Machine with Radial Basis Function (ISVM-RBF), Decision Trees (DT), and K-Nearest Neighbors (KNN) techniques.

Some of the researchers proposed various ensemble learning model to enhance the accuracy of the DR image classification. Antal \textit{et al.} \cite{antal2014ensemble} classifies images based on characteristic features extracted by lesion detection and anatomical part recognition algorithms and achieved $90\%$ 0f accuracy in Messidor database. Somasundaram \textit{et al.} \cite{sk2017machine} showed that the  ensamble machine learning model perform better than the machine learning model alone. Sikder \textit{et al.} \cite{sikder2021severity} employed ensemble learning with feature extraction via histogram and GLCM. Utilizing XGBoost, then Genetic Algorithm, we achieved $94.20\%$ classification accuracy by identifying relevant features. ensemble of five deep CNN models trained on Kaggle dataset. Qummar \textit{et al.} \cite{qummar2019deep} demonstrate improved detection across all DR stages, achieving better classification accuracy at all the levels (0 to 4: 0—no DR, 1—mild, 2—moderate, 3—severe, 4—indicates proliferative). The authors have improved the prediction accuracy but they have not considered the multi labeling of the dataset. In this paper, the labels are converted into multi labels and a balanced class of the dataset is created at the time of data preprocessing. Decisions from multiple base models; Densenet121 and InceptionV3, are combined to improve the overall performance of the model. 

The key offerings of the manuscript are as follows:
\begin{enumerate}[1.]
    \item Propose an ensemble learning technique by considering two base models; Densenet121 and InceptionV3, for early detection and management of diabetic retinopathy.
    \item Generate a balanced dataset by oversampling of minority classes using the multi label formatting is the data preprocessing phase. 
    \item Evaluate the performance of the proposed ensemble model on the validation set of the dataset. 
    \item Provide a comparative analysis with the previously existing models.
\end{enumerate}

The subsequent sections of the paper are structured as follows: Section \ref{sec2} discusses the related deep learning models comprehensively. In Section \ref{sec3}, we delve into the proposed strategies which are divided into multiple phases. The results and in-depth analysis are presented in Section \ref{sec4}. Finally, Section \ref{sec5} provides the conclusions drawn from the proposed research and outlines the future direction.

\section{Related Work} \label{sec2}
When it comes to diabetic retinopathy (DR) image classification, several Convolutional Neural Network (CNN) models have been employed to achieve accurate results. In this section, some of the notable CNN architectures are discussed which are used in DR classification. 

\subsection*{Visual Geometry Group 16 (VGG16)}
VGG16 \cite{Rocha2022DiabeticRC} is a deep CNN architecture with 16 weight layers, comprising a series of convolutional, pooling, and fully connected layers. It utilizes small $3 \times 3$ convolutional filters to capture fine details and complex patterns in images. It is known for its uniform and simple structure and it performs well on a variety of image classification tasks, including DR classification. However, its deep structure can lead to higher computational costs, making it less efficient for resource-limited environments.

\subsection*{Inception V3 (GoogLeNet)}
Inception V3\cite{inceptionvx}, also known as GoogLeNet, employs inception modules that use multiple filter sizes in parallel within a single layer to capture various feature scales. The use of inception modules improves feature extraction while keeping the model computationally efficient by reducing the number of parameters. It provides high accuracy in DR classification, due to its ability to capture a wide range of features.

\subsection*{Xception}
Xception \cite{Xception} extends the Inception architecture by replacing standard convolutions with depthwise separable convolutions. This modification reduces the number of parameters and computational cost while maintaining or even improving performance. It is well suited in DR classification as it can efficiently extract complex features from images. 

\subsection*{DenseNet121}
DenseNet121 \cite{zhang2021classification} is a CNN with $121$ layers featuring dense connectivity, where each layer is connected to every other layer in a feed-forward manner. This dense connectivity allows for better feature reuse and efficient gradient flow throughout the network. This helps the model to learn and leverage complex features from images leading to strong performance on image classification tasks especially DR classification.

\subsection*{Ensemble model}

Ensemble models have gained attention for their effectiveness in predicting diabetic retinopathy due to their ability to combine multiple deep learning architectures for improved performance. Chaurasia \textit{et al.} \cite{Chaurasia2023TransferLE} proposed a transfer learning-driven ensemble model that integrates six CNN models, including DenseNet-169, VGG-19, ResNet101-V2, MobileNet-V2, and Inception-V3, achieving up to $98\%$ accuracy in detecting DR stages. Another research \cite{qummar2019deep} used an ensemble approach with publicly available retinal image datasets to train five deep CNN models (ResNet50, InceptionV3, Xception, DenseNet121, DenseNet169) to encode rich features and enhance classification across different stages of DR. Moreover, Kale \textit{et al.} \cite{kale}  proposed a stacked ensemble model utilized to classify five severity levels of DR (ranging from No-DR to proliferative-DR), which improved accuracy in identifying the correct stage of the disease. 

In addition to these models, \textit{transfer learning} \cite{inceptionvx, kale} has been a key strategy in DR classification. By fine-tuning pre-trained models such as VGG16, Inception V3, and DenseNet121 on specific DR datasets, researchers have been able to achieve high accuracy and robust performance even with smaller training datasets. These models leverage rich features learned from large datasets, predictive capabilities of deep learning approaches and can be adapted effectively to new tasks such as DR classification and detection. 

\section{Proposed Approach} \label{sec3}
In this section, the approaches to the design of the ensemble model is discussed.  The section contains architectural diagram, training details, hyper-parameters, data preprocessing steps, transfer learning mechanism with CNN, resampling and, multi label approach and evaluation and refinement parameters. 

\subsection{Data Acquisition and Preprocessing}
\begin{figure}[htb!]
    \centering
    \begin{subfigure}[b]{0.3\textwidth}
        \includegraphics[width=\textwidth]{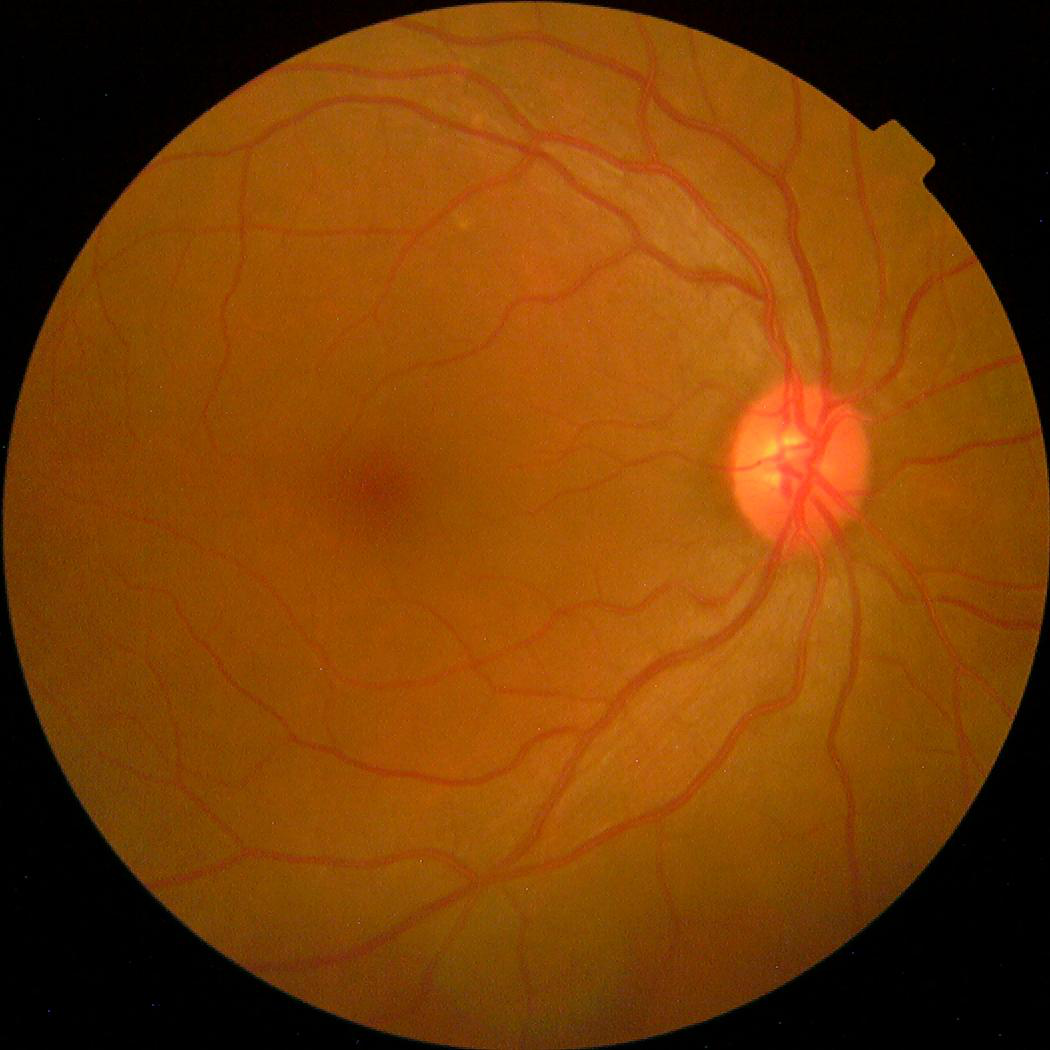}
        \caption{Image 1}
        \label{fig:preprocessing_1}
    \end{subfigure}
    \hfill
    \begin{subfigure}[b]{0.3\textwidth}
        \includegraphics[width=\textwidth]{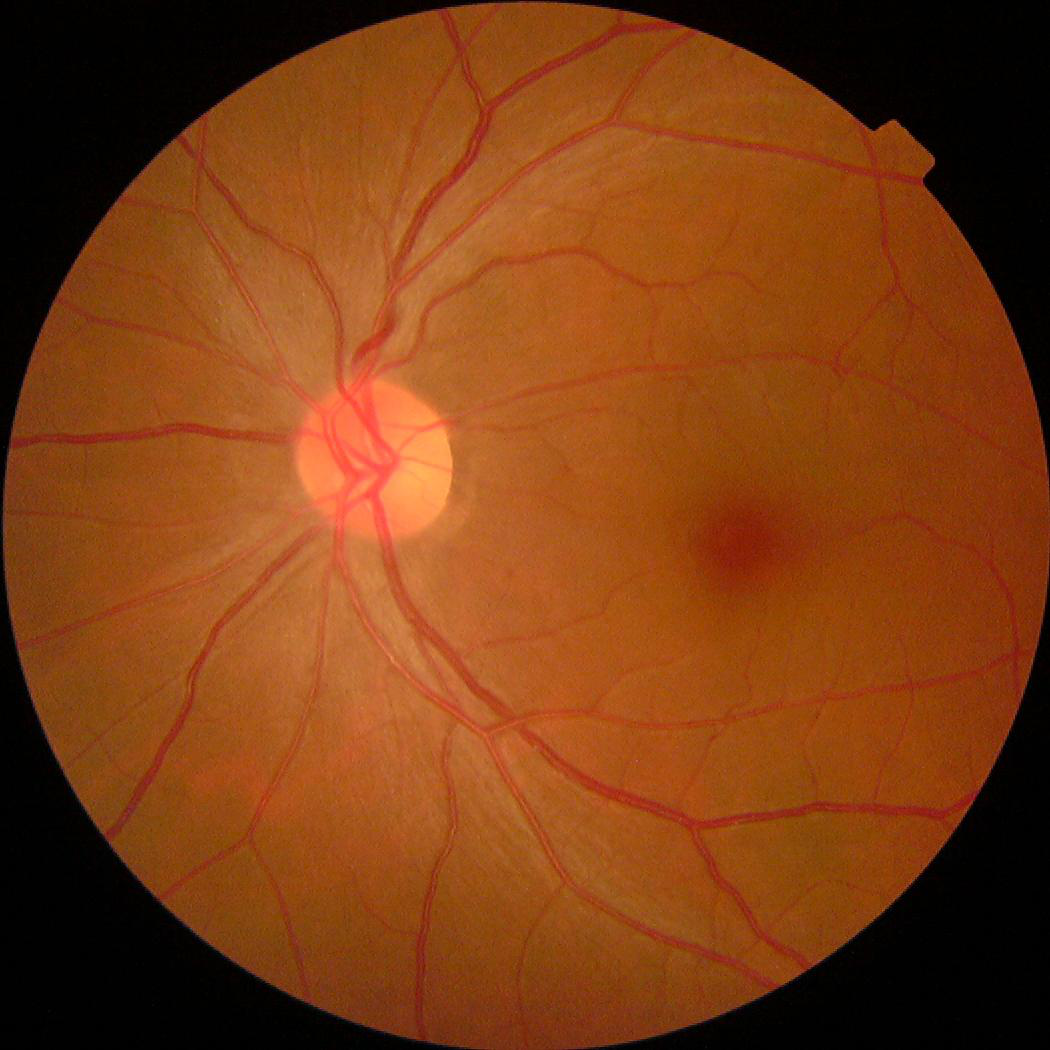}
        \caption{Image 2}
        \label{fig:preprocessing_2}
    \end{subfigure}
    \hfill
    \begin{subfigure}[b]{0.3\textwidth}
        \includegraphics[width=\textwidth]{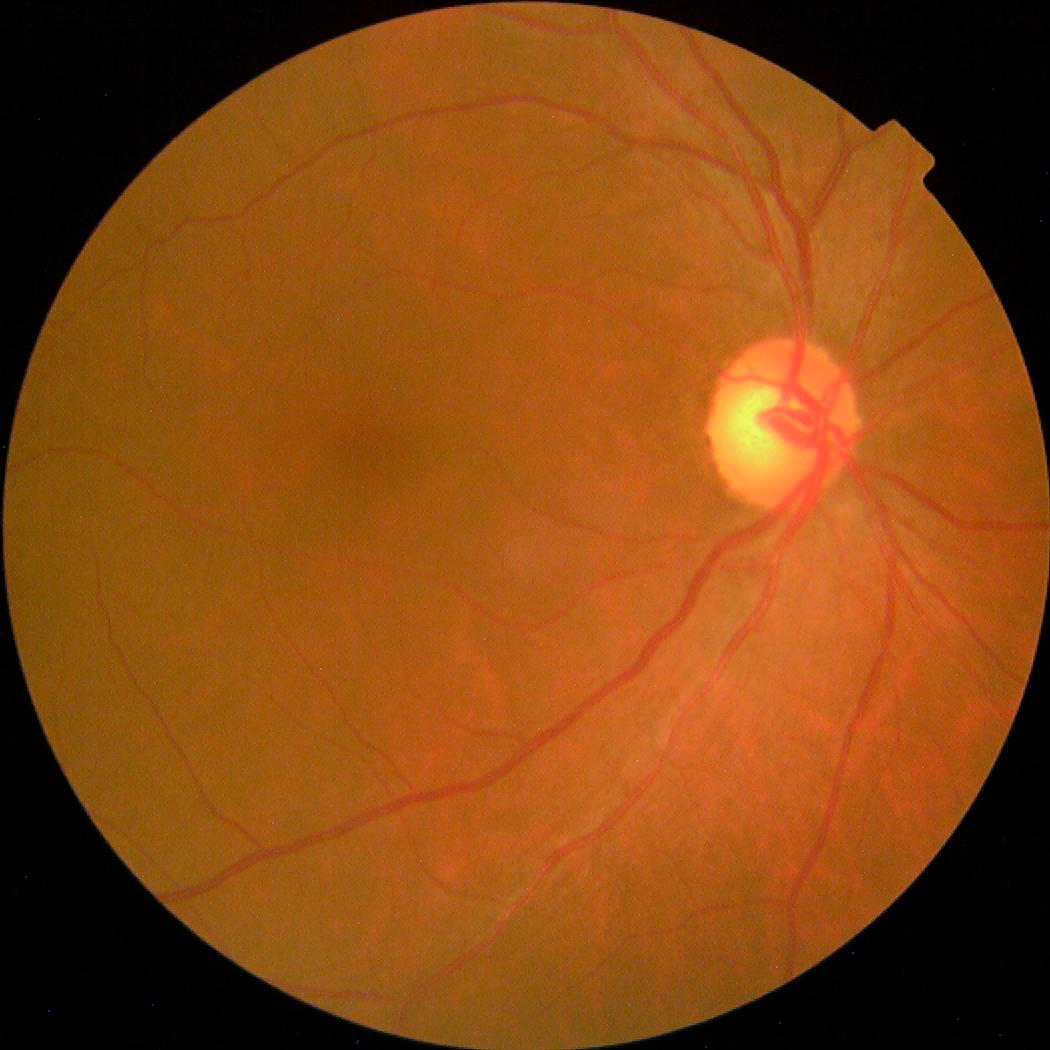}
        \caption{Image 3}
        \label{fig:preprocessing_3}
    \end{subfigure}
    \caption{Dataset Before Preprocessing}
    \label{fig:preprocessing}
\end{figure}

\begin{figure}[htb!]
    \centering
    \begin{subfigure}[b]{0.3\textwidth}
        \includegraphics[width=\textwidth]{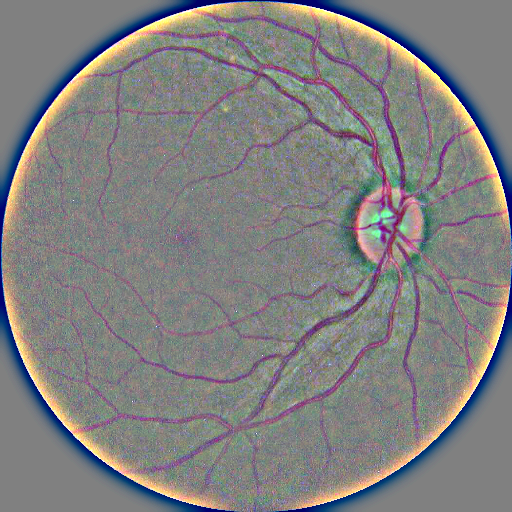}
        \caption{Image 1}
        \label{fig:preprocessing_1}
    \end{subfigure}
    \hfill
    \begin{subfigure}[b]{0.3\textwidth}
        \includegraphics[width=\textwidth]{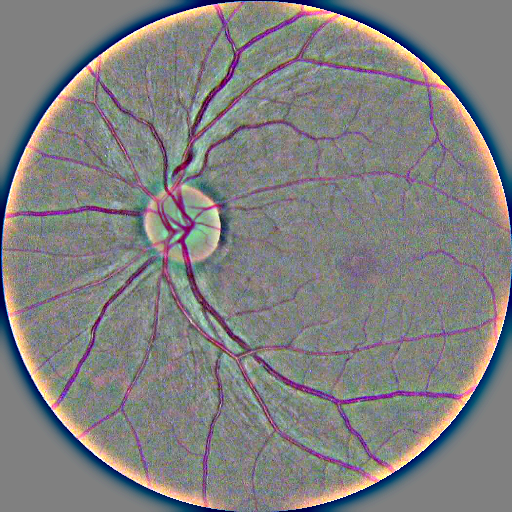 }
        \caption{Image 2}
        \label{fig:preprocessing_2}
    \end{subfigure}
    \hfill
    \begin{subfigure}[b]{0.3\textwidth}
        \includegraphics[width=\textwidth]{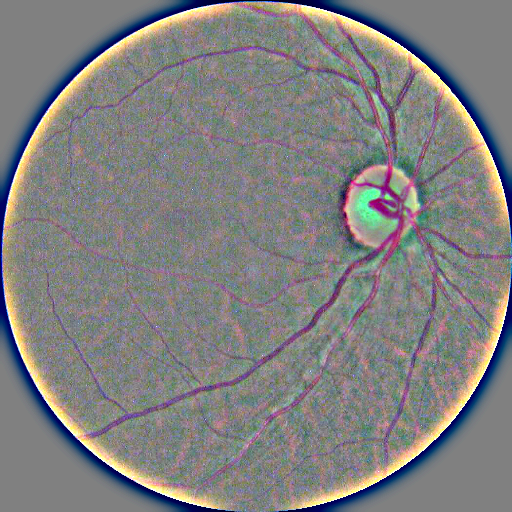}
        \caption{Image 3}
        \label{fig:preprocessing_3}
    \end{subfigure}
    \caption{Dataset After Preprocessing}
    \label{fig:afterpreprocessing}
\end{figure}

A public dataset, APTOS 2019 \cite{aptos2019} of retinal images labeled with the corresponding eye disease is chosen for the prediction of diabetic retinopathy. Some of the images of the datasets before applying preprocessing is shown in Fig. \ref{fig:preprocessing}. The preprocessing
of retinal images helps to improve model performance and accuracy. Hence, some parameters such as random horizontal and vertical flips,
random zooming (zoom range: 0.15), 
constant fill mode for points outside image boundaries, are considered for preprocessing the images. Some of the preprocessed images are shown in Fig. \ref{fig:afterpreprocessing}. The steps to perform preprocessing is as follows:

\begin{itemize}
    \item \textbf{Step 1:} Crop any extra dark pixels present at the sides of the images.
    \item \textbf{Step 2:} Standardize the image size by resizing it.
    \item \textbf{Step 3:} Draw a circle from the center of the image to provide a consistent shape for all images.
    \item \textbf{Step 4:} Apply a smoothing technique using Gaussian blur to remove noise caused by varying resolutions and lighting conditions.
\end{itemize}

The formulation of the output image is provided in Eq. \eqref{Eq1}.
\begin{equation}
    \text{output}(x, y) = \frac{1}{{2\pi\sigma_x\sigma_y}} \cdot \sum_{i=-k}^{k} \sum_{j=-k}^{k} \text{input}(x+i, y+j) \cdot \exp\left(-\frac{i^2}{2\sigma_x^2} - \frac{j^2}{2\sigma_y^2}\right). \label{Eq1}
\end{equation}
Where \(\text{output}(x, y)\) is the value of the pixel at coordinates \((x, y)\) in the output image, \(\text{input}(x+i, y+j)\) is the value of the pixel at coordinates \((x+i, y+j)\) in the input image, \(\sigma_x\) and \(\sigma_y\) are the standard deviations in the X and Y directions, which determine the amount of blur, and \(k\) is the half-size of the kernel in both the X and Y directions.

\subsection{Convolutional Neural Network (CNN) }

Due to multi-stage architecture, CNN is the most efficient deep learning model for classifying images data. It consists of several layers, including convolution (CONV), activation (ReLU), and pooling (POOL). Each of these layers transforms the input data in specific ways. Here are the steps in CNN \cite{cnn}:

\begin{enumerate}[1.]
    \item  \textbf{Convolution (CONV) Layer:} It accepts an input volume of size $[X_1 \times Y_1 \times Z_1]$, where $X_1$ is the width, $Y_1$ is the height, and $Z_1$ is the depth. The output volumes (i.e., convolution maps) are of size $[X_2 \times Y_2 \times Z_2]$, is calculated using the following equations:
        
        \[
        X_2 = \left\lfloor \frac{X_1 + 2 \cdot P - F}{S} \right\rfloor + 1
        \]
        \[
        Y_2 = \left\lfloor \frac{Y_1 + 2 \cdot P - F}{S} \right\rfloor + 1
        \]
        \[
        Z_2 = K
        \]
        where, $F$ is the spatial extent of the filter, $K$ is the number of filters or convolution kernels, $P$ is the amount of zero padding, and $S$ is the stride (step size with which the filter slides).
    \item \textbf{Activation (ReLU) Layer:} This layer applies an activation function (such as the max function) element-wise to introduce non-linearity to the output of the CONV layer. The function commonly used is ReLU: $f(x) = \max(0, x)$. This operation does not change the size of the volume.
    \item \textbf{Pooling (POOL) Layer:} It uses the MAX operation to optimize the spatial size of the representation and reduce the number of parameters. The POOL layer produces an output volume of size $[X_2 \times Y_2 \times Z_2]$ and is calculated using:

        \[
        X_2 = \left\lfloor \frac{X_1 - F}{S} \right\rfloor + 1
        \]
        \[
        Y_2 = \left\lfloor \frac{Y_1 - F}{S} \right\rfloor + 1
        \]
        \[
        Z_2 = Z_1
        \]
        Where $F$ is the size of the pooling region and $S$ is the stride.
    \item \textbf{Flattening and Fully Connected (FC) Layers:} After the feature extraction, the output from the POOL layer is flattened into a feature vector. This vector is then connected to a series of fully connected layers (multilayer perceptron) to perform classification on the features. 
\end{enumerate}
These steps combine to form the feature extraction and classification process in a CNN.

\subsection{Transfer Learning with CNNs}
Select a pre-trained CNN model such as InceptionV3, DenseNet121, or EfficientNet, which are commonly trained on extensive image datasets like ImageNet. Preserve the initial layers of the pre-trained model, as they are responsible for extracting general image features and should remain unchanged. Add new layers on top of the frozen pre-trained model. These new layers will be specific to the eye disease classification task. The number and type of layers will depend on the complexity of the problem. 

\subsection{Resampling}
A resampling technique is proposed for addressing class imbalance in multi-label classification tasks. By setting a desired number of samples per class, we ensured balanced representation across all classes in the dataset. Our approach involved replicating indices of samples for each class based on the desired count which is 700, with additional random sampling to account for any remainder. Through this process, we aimed to mitigate the biases introduced by class imbalance, fostering the development of more equitable and accurate multi-label classification models. Before resampling, our dataset exhibited class imbalances: $1543$ samples in class $0$, $314$ samples in class $1$, $849$ samples in class $2$, $164$ samples in class $3$, and $251$ samples in class $4$. However, after resampling, all classes contained precisely 700 samples. 

\subsection{Multi Label}

In the proposed work, the target encoding approach is modified for the transition from a single-label to multilabel classification. Traditionally, a class label after one hot encoding would be represented by a single $"1"$ in a binary vector, with other classes indicated by $"0"$s. However, we redefined this scheme to signify the inclusion of all preceding classes up to the target class. For example, instead of encoding class $4$ retinopathy as $[0, 0, 0, 1]$, the proposed approach predicts $[1, 1, 1, 1]$, encompassing all prior classes. This adaptation acknowledges the progressive nature of retinopathy severity, by enhancing the model's capacity to differentiate variations in disease progression. Detailed steps are provided in Flow diagram in Figure \ref{fig:methodflow}.

\subsection{Architecture of the proposed ensemble model} 
 
% Flow  
% Data Preprocessing

% Resize all images to 224x224 pixels
% Convert labels to multi-label format (e.g., [0, 0, 0, 1] becomes [1, 1, 1, 1])
% Oversample the minority classes to create a balanced dataset

% Base Model Training
% Define and compile two base models: DenseNet121 and InceptionV3
% Train both base models on the dataset
% Monitor validation quadratic weighted kappa during training
% Stacking Ensemble
% Combine the predictions from both base models by stacking them as input features
% Define and compile the meta model
% Train the meta-model on the stacked features from base models
% Monitor validation quadratic weighted kappa during training
% Ensemble Evaluation
% Evaluate the performance of the stacking ensemble on the validation set
The process for predicting DR disease is outlined in Fig. \ref{fig:methodflow}. Initially, data preprocessing is performed to enhance accuracy. Subsequently, DenseNet121 and InceptionV3 base models are trained on the APTOS dataset. In the third step, the outcomes from both base models are aggregated and further trained. Ultimately, the prediction score is assessed.
\usetikzlibrary{shapes, arrows, positioning}
\begin{figure}
\centering
\begin{tikzpicture}[node distance=1.2cm]

    % Define styles for blocks and lines
    \tikzstyle{process} = [rectangle, draw, fill=blue!20, text width=28em, text centered, rounded corners, minimum height=4em]
    \tikzstyle{arrow} = [draw, -latex, thick]
    
    % Data Preprocessing
    \node[process] (preprocessing) {
        Data Preprocessing
        \begin{itemize}
            \item Resize all images to $224\times224$ pixels
            \item Convert labels to multi-label format (e.g., $[0, 0, 0, 1]$ becomes $[1, 1, 1, 1]$)
            \item Oversample the minority classes to create a balanced dataset
        \end{itemize}
    };

    % Base Model Training
    \node[process, below=of preprocessing] (training) {
        Base Model Training
        \begin{itemize}
            \item Define and compile two base models: DenseNet121 and InceptionV3
            \item Train both base models on the dataset
            \item Monitor validation quadratic weighted kappa during training
        \end{itemize}
    };

    % Stacking Ensemble
    \node[process, below=of training] (stacking) {
        Stacking Ensemble
        \begin{itemize}
            \item Combine the predictions from both base models by stacking them as input features
            \item Define and compile the meta model
            \item Train the meta-model on the stacked features from base models
            \item Monitor validation quadratic weighted kappa during training
        \end{itemize}
    };

    % Ensemble Evaluation
    \node[process, below=of stacking] (evaluation) {
        Ensemble Evaluation
        \begin{itemize}
            \item Evaluate the performance of the stacking ensemble on the validation set
        \end{itemize}
        
    };

    % Connect the processes with arrows
    \path [arrow] (preprocessing) -- (training);
    \path [arrow] (training) -- (stacking);
    \path [arrow] (stacking) -- (evaluation);
    \label{methodflow}
\end{tikzpicture}
\caption{Flow diagram of the steps involved from data pre-processing to Ensemble model evaluation of the diabetic retinopathy disease.} \label{fig:methodflow}
\end{figure}
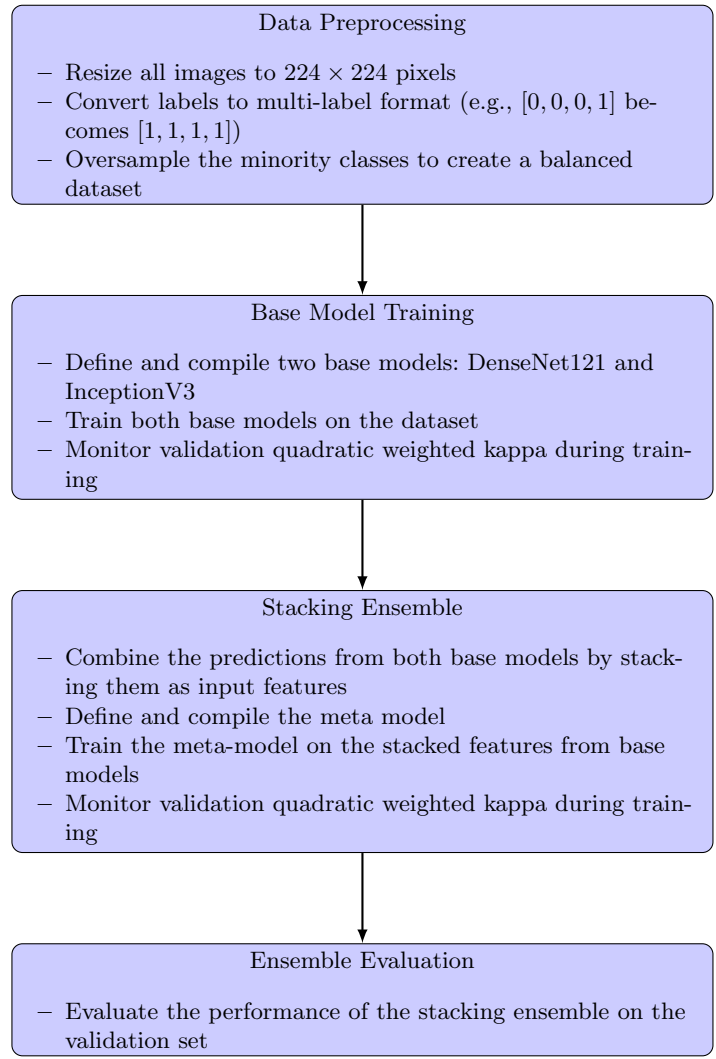

\begin{figure}[htb!]
    \centering
    \includegraphics[width=0.9\textwidth]{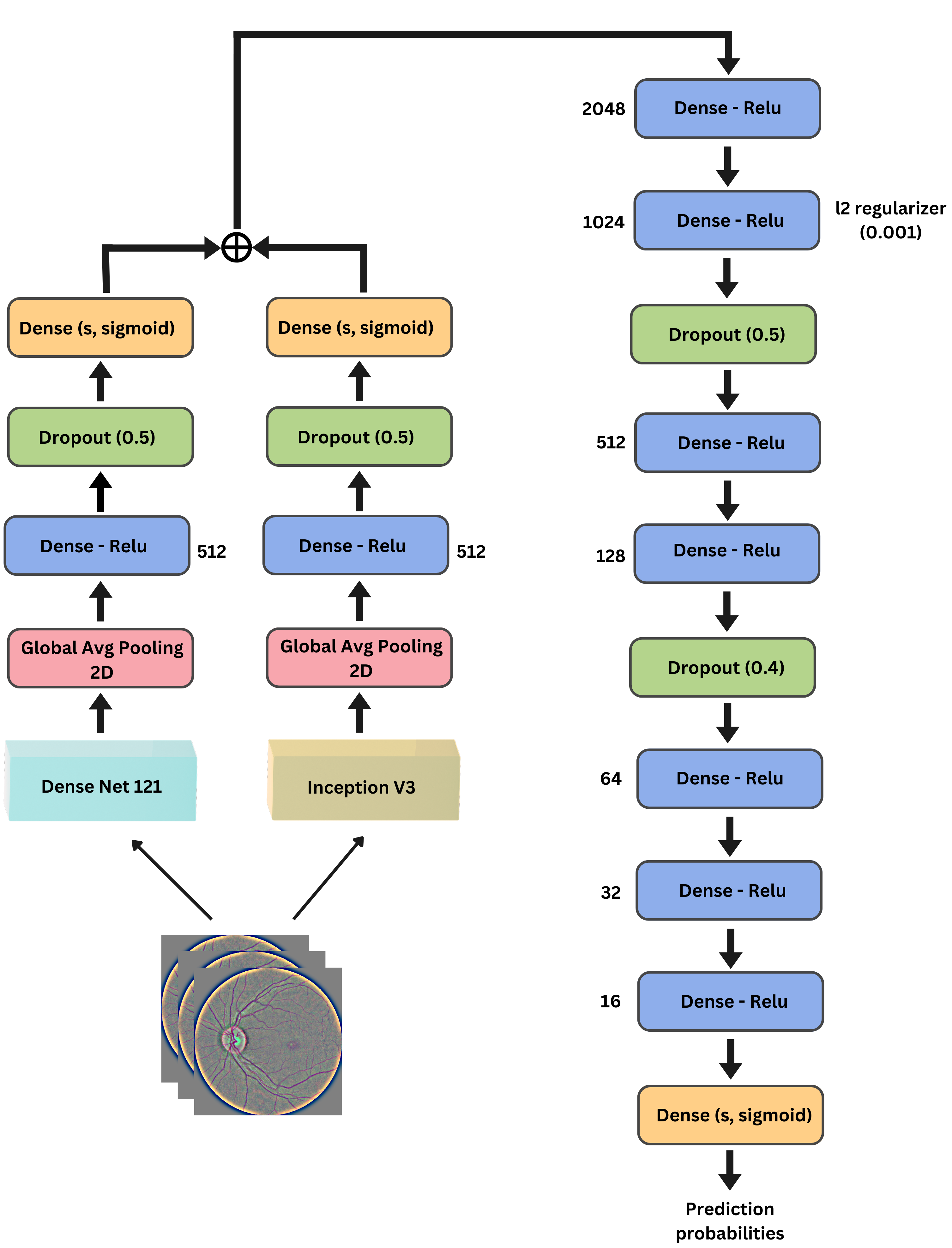}
    \caption{Model Architceture}
    \label{fig:arch}
\end{figure}

The architectural diagram of the proposed model is shown in Fig. \ref{fig:arch}. Transfer learning was employed using pre-trained models, namely DenseNet121 \cite{densenet} and InceptionV3 \cite{inceptionvx}. The models were fine-tuned on the retinal image dataset using a custom multi-label classification approach. The target variable was transformed into a multi-label format, where each class encompasses all the classes before it. 

The model consists of two parts. The first part uses DenseNet121 as the base model, followed by a sequence of layers: pooling, dense (ReLU activation), dropout, and sigmoid activation. The second part mirrors the first, but with InceptionV3 as the base model. The outputs from both parts are combined and passed through alternating dense (ReLU activation) and dropout layers, ending with a sigmoid activation function for the final output. The model's loss is calculated using Binary Cross Entropy between the predicted output and the true labels. The steps are provided here in \ref{steps}.

\begin{minipage}[t]{0.45\textwidth}
\centering
\[
\begin{aligned}
y_{11} & = \text{densenet121} - (1) \\
y_{12} & = \text{pooling}(y_{11}) - (2) \\
y_{13} & = \text{dense\_relu}(y_{12}) - (3) \\
y_{14} & = \text{dropout}(y_{13}) - (4) \\
y_{15} & = \text{sigmoid}(y_{14}) - (5) \\
\\
y_{21} & = \text{inceptionV3} - (6) \\
y_{22} & = \text{pooling}(y_{21}) - (7) \\
y_{23} & = \text{dense\_relu}(y_{22}) - (8) \\
y_{24} & = \text{dropout}(y_{23}) - (9) \\
y_{25} & = \text{sigmoid}(y_{24}) - (10) \\
\\
y_{3} & = \text{stack}(y_{15},y_{25}) - (11)
\end{aligned}
\]
\end{minipage}%
\hfill
\begin{minipage}[t]{0.45\textwidth}
\centering
\[
\begin{aligned}
y_{31} & = \text{dense\_relu}(y_{3}) - (12) \\
y_{32} & = \text{dense\_relu}(y_{31}) - (13) \\
y_{33} & = \text{dropout}(y_{32}) - (14) \\
y_{34} & = \text{dense\_relu}(y_{33}) - (15) \\
y_{35} & = \text{dense\_relu}(y_{34}) - (16) \\
y_{36} & = \text{dropout}(y_{35}) - (17) \\
y_{37} & = \text{dense\_relu}(y_{36}) - (18) \\
y_{38} & = \text{dense\_relu}(y_{37}) - (19) \\
y_{39} & = \text{dense\_relu}(y_{38}) - (20) \\
\text{pred}_i & = \text{sigmoid}(y_{39}) - (21)
\end{aligned}
\]
\[
\text{loss} = \text{BCrossEntropy}(\text{pred}_i,y_i)
\]
\label{steps}
\end{minipage}

\subsection{Training details and hyperparameters}

The model training and evaluation process involved the multiple steps. In the data Preprocessing, the images were resized to $224 \times 224$ pixels and preprocessed using the OpenCV and PIL libraries. A balanced dataset was created by oversampling the minority classes to a desired number of samples (500 per class) to handle class imbalance. The specification for the hyperparameters are provided in Table \ref{tab:hyperparameter}.

\begin{table}[]
    \centering
    \caption{Hyperparameters}
    \begin{tabular}{ll}
    \hline
    Hyper parameters & Specification \\ \hline
     Loss Function : & Binary Cross-Entropy  \\
     Optimizer : & Adam with a learning rate of 0.00005 \\
     Batch Size : & 32 \\
     Regularization: & L2 (0.001) on Dense layers \\
     Dropout Rate: & 0.5 \\
    Data Augmentation: & Random zoom(0.15), horizontal flip, and vertical flip \\
    Epochs: & $15$ epochs for DenseNet121 and InceptionV3 models \\ \hline
    \end{tabular}
    \label{tab:hyperparameter}
\end{table}

After training the individual models, a stacking ensemble approach was used by combining the predictions from DenseNet121 and InceptionV3 models. A meta-model was trained on these stacked predictions to obtain the final predictions. A neural network with multiple dense layers and dropout regularization are used. Total 200 epochs, batch size of 64, binary cross-entropy loss, and Adam optimizer. Finally the models were evaluated using precision, recall, F1-score, quadratic weighted Kappa score, and accuracy.

\subsection{Evaluation and Refinement}

The performance of the fine-tuned CNN models are evaluated on a separate validation dataset (not used for training). Metrics like accuracy, precision, recall, F1 score, and confusion matrix can be used. The definition of all these parameters are provided as follows:
\begin{itemize}
    \item \textbf{Accuracy:} Measures the proportion of correctly predicted outcomes over the total number of predictions made
    \item \textbf{Recall: } It is defined as the number of positive samples the model correctly recalled out of the total number of actual positive samples (TP + FN).
    \item \textbf{Precision:} It is defined as the number of positive samples that the model was able to precisely predict out of the total number of positive predicted samples (TP + FP).
    \item \textbf{F1 score:} It is a single metric that provides a way to combine both precision and recall into one score by taking harmonic mean of precision and recall.
    \begin{center}
     $F_1 = \frac{2 \times \text{precision} \times \text{recall}}{\text{precision} + \text{recall}}$
    \end{center}
    \item \textbf{Confusion matrix:} It compares model predictions against actual labels to compute metrics like accuracy, precision, recall, and F1 score, providing insights into classification performance for multi-class problems. The nearer the value is to $1$, the more accurate and reliable the model's predictions become.
\end{itemize}

\section{Results and Analysis} \label{sec4}
In this section, DR prediction accuracy is compared with some well-known machine learning and deep learning models, and performance is evaluated considering accuracy, precision, recall, and F1 score. A confusion matrix is also plotted for the models. 

\subsection{Dataset description}

This study utilizes the publicly available APTOS dataset for diabetic retinopathy detection, which consists of fundus images. These images originate from various camera models, potentially leading to differences in visual appearance between left and right images. The dataset categorizes the sensitivity of diabetic retinopathy presence on a scale of  0 to 4: 0—no DR, 1—mild, 2—moderate, 3—severe, 4—indicates proliferative DR. From the dataset, a random sample of 5,590 images across five categories is selected. Specifically, 3,662 images (3,112 for training and 1,928 for testing) are employed. Prior to training, the dataset undergoes preprocessing steps to ensure its quality and suitability for the task at hand. 

\subsection {Evaluation of the proposed ensemble model on the
dataset}

\begin{figure}[htb!]
    \centering
    \begin{subfigure}[b]{0.45\textwidth}
        \includegraphics[width=\textwidth]{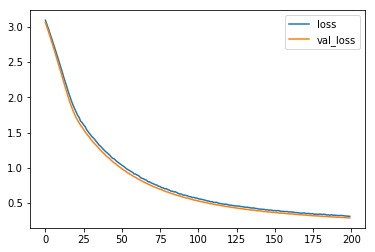}
        \caption{Loss Curve}
        \label{fig:loss_curve}
    \end{subfigure}
    \hfill
    \begin{subfigure}[b]{0.45\textwidth}
        \includegraphics[width=\textwidth]{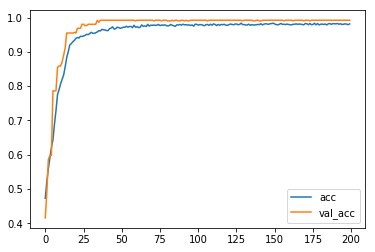}
        \caption{Accuracy Curve}
        \label{fig:accurey_curve}
    \end{subfigure}
    \caption{Validation loss and validation Accuracy Curve}
    \label{fig:lossacc}
\end{figure}

During both the training and validation phases, the performance of the proposed ensemble model is assessed by plotting validation loss (in Fig. \ref{fig:lossacc}(a)) and accuracy curves (in Fig. \ref{fig:lossacc}(b)). Validation loss serves as an indicator of the model's error minimization on unseen data, while validation accuracy reflects its predictive capability. From the figures, it can be analyzed that the validation loss and accuracy is improving over epochs.  Moreover, the gap between training and validation metrics suggests that the model is not overfitting to the training data.

\subsection{Discussion of results}

\begin{figure}[htb!]
  \centering
  \includegraphics[width=\linewidth,height=2.5 in]{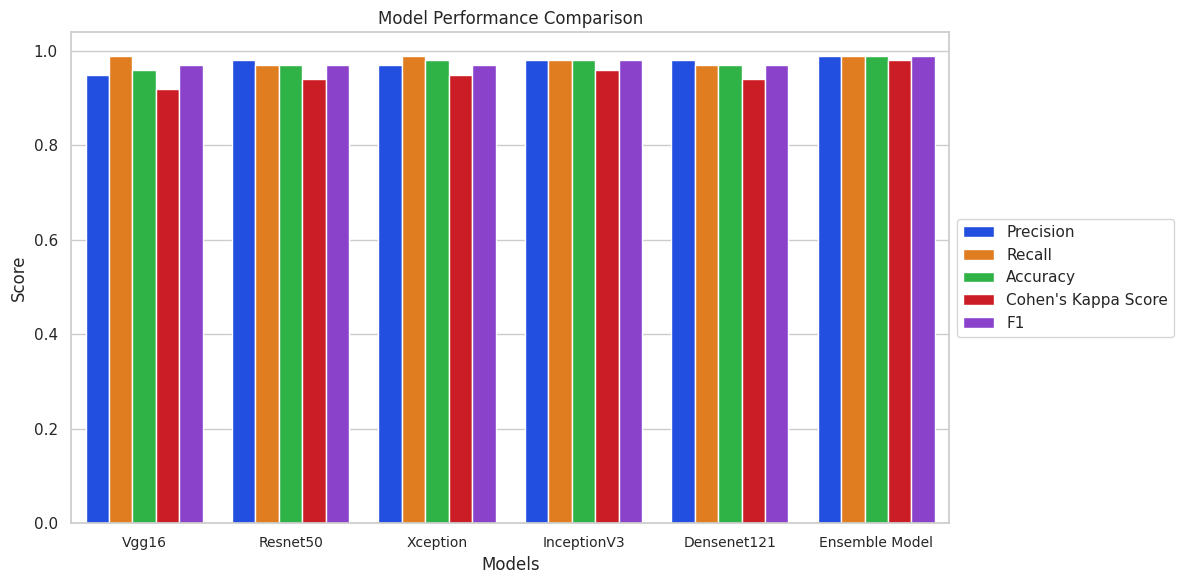}
  \caption{Model Perfomance}
  \label{fig:Model_Perfomance}
\end{figure}

Tabel \ref{tab:my-table} and Fig. \ref{fig:Model_Perfomance} illustrate the superior performance of the proposed hybrid model against other well-known deep learning architectures in detecting diabetic retinopathy. Our model demonstrates higher accuracy and recall, showcasing its effectiveness in this task. Specifically, our ensemble model, leveraging InceptionV3 and Densenet121, achieves outstanding results across all metrics. Notably, it achieves a specificity and precision of 0.99, an accuracy rate of 0.99, and a Cohen's Kappa Score of 0.98. These outcomes underscore the capability of our model to capture both local and global features in retinal images, thereby significantly enhancing diabetic retinopathy detection. Consequently, our ensemble model emerges as a promising tool to support clinicians in diagnosis.

\begin{table}[htb!]
\centering
\caption{Comparative analysis among the proposed model and existing models across evaluation metrics including Precision, Recall, Accuracy, Cohen's Kappa Score, and F1 Score.}
\label{tab:my-table}
\resizebox{\textwidth}{!}{%
\begin{tabular}{@{}llllll@{}}
\toprule
Model                         & Precision & Recall & Accuracy & Cohen's & F1   \\ 
&&& & Kappa Score & \\\midrule

Vgg16\cite{Rocha2022DiabeticRC}     & 0.95      & 0.99   & 0.96     & 0.92                & 0.97 \\
Resnet50     & 0.98      & 0.97   & 0.97     & 0.94           & 0.97 \\
Xception\cite{Xception}  & 0.97      & 0.99   & 0.98     & 0.95             & 0.97 \\
InceptionV3\cite{inceptionvx}       & 0.98      & 0.98   & 0.98     & 0.96        & 0.98 \\
Densenet121\cite{zhang2021classification}       & 0.98      & 0.97   & 0.97     & 0.94         & 0.97 \\
\textbf{Proposed Model}                  & \textbf{0.99}      & \textbf{0.99}   & \textbf{0.99}     & \textbf{0.98 }               & \textbf{0.99} \\ \bottomrule
\end{tabular}%
}
\end{table}

\begin{figure}[htbp]
    \centering
    \begin{subfigure}[b]{\textwidth}
        \includegraphics[width=\textwidth]{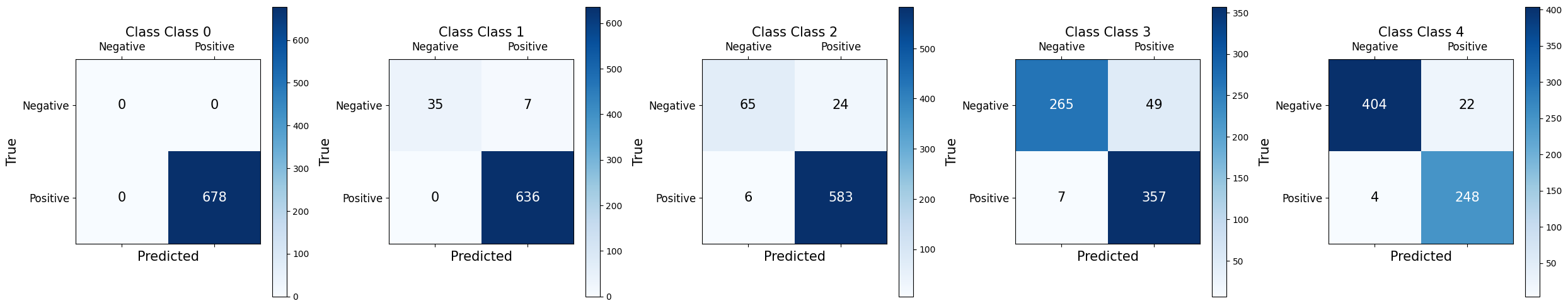}
        \caption{Vgg16}
        \label{fig:vgg16_cm}
    \end{subfigure}
    \hfill
    \begin{subfigure}[b]{\textwidth}
        \includegraphics[width=\textwidth]{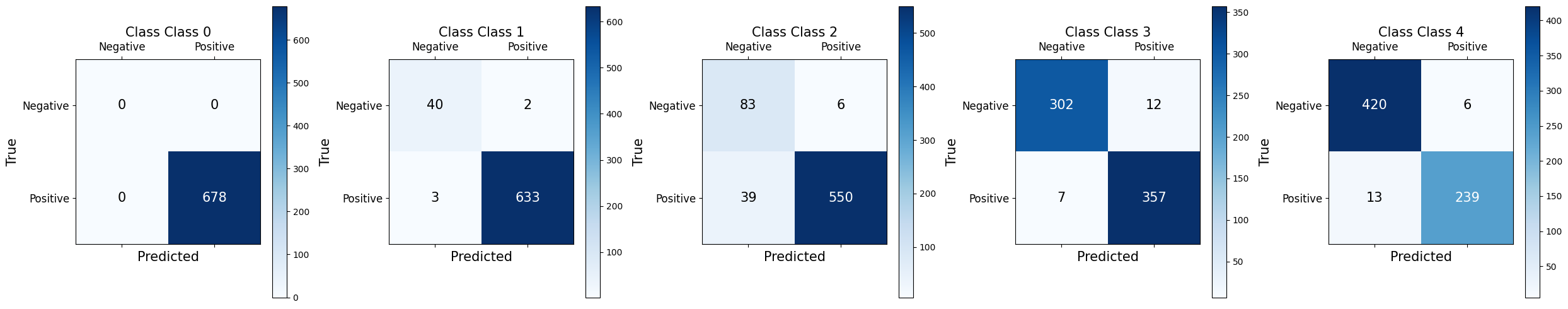}
        \caption{Resnet50}
        \label{fig:Resnet50_cm}
    \end{subfigure}
    \hfill
    \begin{subfigure}[b]{\textwidth}
        \includegraphics[width=\textwidth]{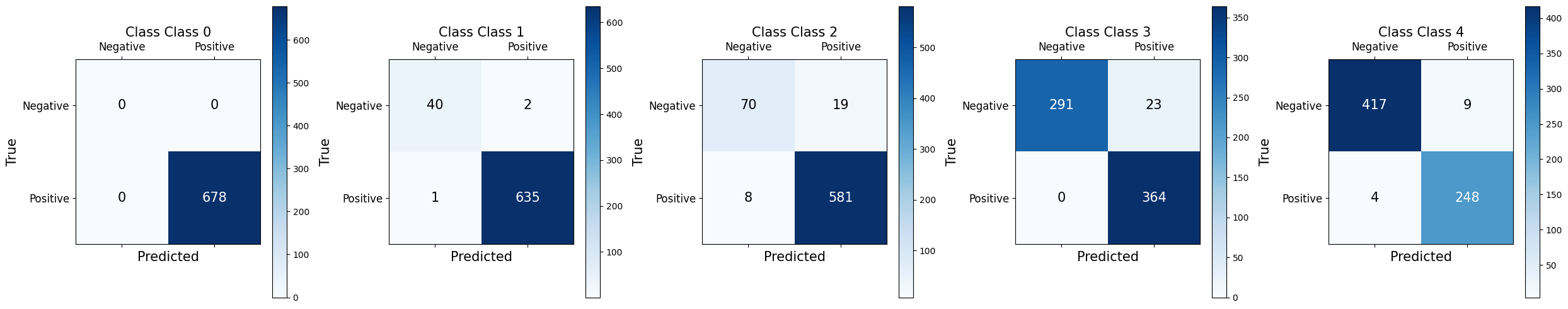}
        \caption{Xception}
        \label{fig:Xeception}
    \end{subfigure}
    \hfill
    \begin{subfigure}[b]{\textwidth}
        \includegraphics[width=\textwidth]{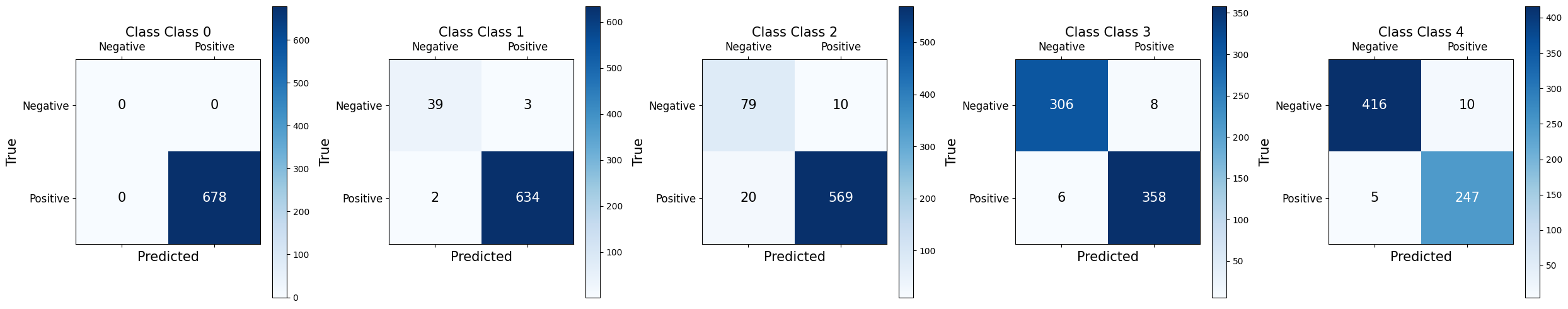}
        \caption{InceptionV3}
        \label{fig:inception_cm}
    \end{subfigure}
    \hfill
    \begin{subfigure}[b]{\textwidth}
        \includegraphics[width=\textwidth]{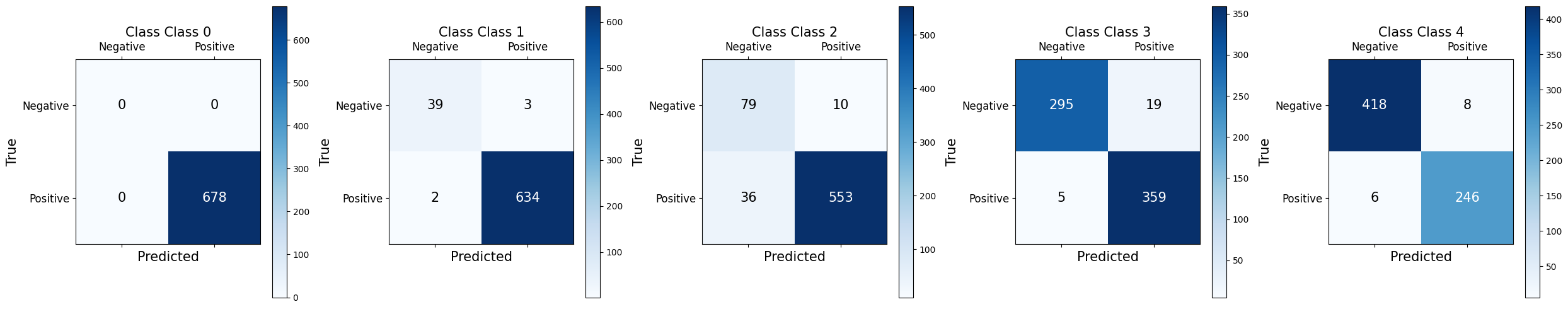}
        \caption{DenseNet121}
        \label{fig:densenet_cm}
    \end{subfigure}
    \hfill
    \begin{subfigure}[b]{\textwidth}
        \includegraphics[width=\textwidth]{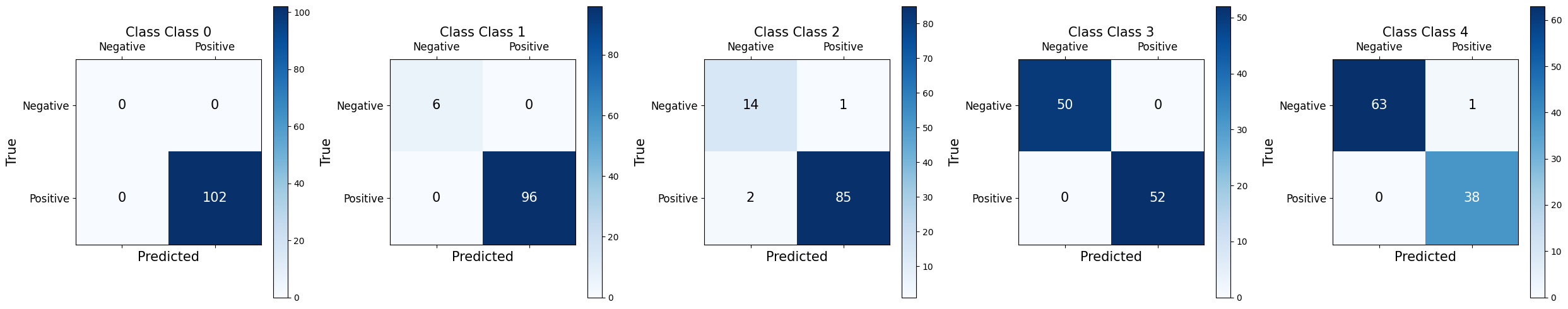}
        \caption{Proposed Ensemble Model}
        \label{fig:ensemble_cm}
    \end{subfigure}
    
    \caption{Confusion Matrices for Various Models}
    \label{fig:confusion_matrices}
\end{figure}

A confusion matrix Fig. \ref{fig:confusion_matrices} is a key tool for evaluating classification models, detailing predictions versus actual labels. It reveals true positives, true negatives, false positives, and false negatives, aiding in accuracy, precision, recall, and F1-score calculations. Insights gained help identify misclassification trends, guiding data gathering and model adjustments for improved performance. 

Furthermore, our proposed model outperforms other hybrid approaches across all metrics, including accuracy, precision, recall, F1 score, and Cohen's Kappa score. This underscores the efficacy of combining InceptionV3 and Densenet121, showcasing a robust strategy for improving performance in diabetic retinopathy detection.

\section{Conclusion and Future Scope} \label{sec5}
In this paper, an ensemble learning-based approach is provided by employing two base models; Densenet121 and Inceptionv3, to enhance the accuracy and early detection capabilities of diabetic retinopathy. During the data preprocessing phase, images are resized to $224 \times 224$ pixels, and a multi-label formatting technique is adopted to establish a balanced dataset. The proposed model undergoes training and testing using the APTOS dataset of fundus images. Comparative analysis is conducted with five other models; namely Vgg16, Resnet50, Xception, InceptionV3, and Densenet121. Results reveal that the proposed model's validation accuracy is $99\%$, Kohen's Kappa score is $98\%$ and F1 score is also $99\%$. Additionally, a confusion matrix is generated to show misclassification trends and overall performance.

In the future, the proposed model could undergo testing on diverse datasets and comparisons with a broader array of models. Furthermore, its applicability can extend beyond diabetic retinopathy to encompass other classification tasks.
\section*{Declaration}
The authors declare no competing interests and no finding is granted to this work.
\bibliographystyle{splncs04}
\bibliography{main}

\begin{thebibliography}{10}
\providecommand{\url}[1]{\texttt{#1}}
\providecommand{\urlprefix}{URL }
\providecommand{\doi}[1]{https://doi.org/#1}

\bibitem{ali2023hybrid}
Ali, G., Dastgir, A., Iqbal, M.W., Anwar, M., Faheem, M.: A hybrid convolutional neural network model for automatic diabetic retinopathy classification from fundus images. IEEE Journal of Translational Engineering in Health and Medicine  (2023)

\bibitem{antal2014ensemble}
Antal, B., Hajdu, A.: An ensemble-based system for automatic screening of diabetic retinopathy. Knowledge-based systems  \textbf{60},  20--27 (2014)

\bibitem{bilal2024improved}
Bilal, A., Imran, A., Baig, T.I., Liu, X., Long, H., Alzahrani, A., Shafiq, M.: Improved support vector machine based on cnn-svd for vision-threatening diabetic retinopathy detection and classification. Plos one  \textbf{19}(1),  e0295951 (2024)

\bibitem{Chaurasia2023TransferLE}
Chaurasia, B.K., Raj, H., Rathour, S.S., Singh, P.B.: Transfer learning--driven ensemble model for detection of diabetic retinopathy disease. Medical \& Biological Engineering \& Computing  \textbf{61}(8),  2033--2049 (2023)

\bibitem{Rocha2022DiabeticRC}
Da~Rocha, D.A., Ferreira, F.M.F., Peixoto, Z.M.A.: Diabetic retinopathy classification using vgg16 neural network. Research on Biomedical Engineering  \textbf{38}(2),  761--772 (2022)

\bibitem{fong2003diabetic}
Fong, D.S., Aiello, L., Gardner, T.W., King, G.L., Blankenship, G., Cavallerano, J.D., Ferris~III, F.L., Klein, R., Association, A.D.: Diabetic retinopathy. Diabetes care  \textbf{26}(suppl\_1),  s99--s102 (2003)

\bibitem{Xception}
Ganesh, M., Dulam, S., Venkatasubbu, P.: Diabetic retinopathy diagnosis with inceptionresnetv2, xception, and efficientnetb3. In: Artificial Intelligence and Technologies: Select Proceedings of ICRTAC-AIT 2020, pp. 405--413. Springer (2021)

\bibitem{he2016deep}
He, K., Zhang, X., Ren, S., Sun, J.: Deep residual learning for image recognition. In: Proceedings of the IEEE conference on computer vision and pattern recognition. pp. 770--778 (2016)

\bibitem{densenet}
Huang, G., Liu, Z., Van Der~Maaten, L., Weinberger, K.Q.: Densely connected convolutional networks. In: Proceedings of the IEEE conference on computer vision and pattern recognition. pp. 4700--4708 (2017)

\bibitem{cnn}
Jmour, N., Zayen, S., Abdelkrim, A.: Convolutional neural networks for image classification. In: 2018 international conference on advanced systems and electric technologies (IC\_ASET). pp. 397--402. IEEE (2018)

\bibitem{kale}
Kale, Y., Sharma, S.: Detection of five severity levels of diabetic retinopathy using ensemble deep learning model. Multimedia Tools and Applications  \textbf{82}(12),  19005--19020 (2023)

\bibitem{aptos2019}
Karthik, Maggie, S.D.: Aptos 2019 blindness detection (2019), \url{https://kaggle.com/competitions/aptos2019-blindness-detection}

\bibitem{lam2018automated}
Lam, C., Yi, D., Guo, M., Lindsey, T.: Automated detection of diabetic retinopathy using deep learning. AMIA summits on translational science proceedings  \textbf{2018}, ~147 (2018)

\bibitem{liang2022end}
Liang, N., Yuan, L., Wen, X., Xu, H., Wang, J.: End-to-end retina image synthesis based on cgan using class feature loss and improved retinal detail loss. IEEE Access  \textbf{10},  83125--83137 (2022)

\bibitem{macsik2022local}
Macsik, P., Pavlovicova, J., Goga, J., Kajan, S.: Local binary cnn for diabetic retinopathy classification on fundus images. Acta Polytech. Hung  \textbf{19}(7),  27--45 (2022)

\bibitem{mirza2014conditional}
Mirza, M., Osindero, S.: Conditional generative adversarial nets. arXiv preprint arXiv:1411.1784  (2014)

\bibitem{nahiduzzaman2021hybrid}
Nahiduzzaman, M., Islam, M.R., Islam, S.R., Goni, M.O.F., Anower, M.S., Kwak, K.S.: Hybrid cnn-svd based prominent feature extraction and selection for grading diabetic retinopathy using extreme learning machine algorithm. IEEE Access  \textbf{9},  152261--152274 (2021)

\bibitem{pratt2016convolutional}
Pratt, H., Coenen, F., Broadbent, D.M., Harding, S.P., Zheng, Y.: Convolutional neural networks for diabetic retinopathy. Procedia computer science  \textbf{90},  200--205 (2016)

\bibitem{qummar2019deep}
Qummar, S., Khan, F.G., Shah, S., Khan, A., Shamshirband, S., Rehman, Z.U., Khan, I.A., Jadoon, W.: A deep learning ensemble approach for diabetic retinopathy detection. Ieee Access  \textbf{7},  150530--150539 (2019)

\bibitem{sikder2021severity}
Sikder, N., Masud, M., Bairagi, A.K., Arif, A.S.M., Nahid, A.A., Alhumyani, H.A.: Severity classification of diabetic retinopathy using an ensemble learning algorithm through analyzing retinal images. Symmetry  \textbf{13}(4), ~670 (2021)

\bibitem{sk2017machine}
SK, S., P, A.: A machine learning ensemble classifier for early prediction of diabetic retinopathy. Journal of Medical Systems  \textbf{41},  1--12 (2017)

\bibitem{inceptionvx}
Szegedy, C., Vanhoucke, V., Ioffe, S., Shlens, J., Wojna, Z.: Rethinking the inception architecture for computer vision. In: Proceedings of the IEEE conference on computer vision and pattern recognition. pp. 2818--2826 (2016)

\bibitem{tufail2021diagnosis}
Tufail, A.B., Ullah, I., Khan, W.U., Asif, M., Ahmad, I., Ma, Y.K., Khan, R., Kalimullah, Ali, M.S.: Diagnosis of diabetic retinopathy through retinal fundus images and 3d convolutional neural networks with limited number of samples. Wireless Communications and Mobile Computing  \textbf{2021},  1--15 (2021)

\bibitem{wan2018deep}
Wan, S., Liang, Y., Zhang, Y.: Deep convolutional neural networks for diabetic retinopathy detection by image classification. Computers \& Electrical Engineering  \textbf{72},  274--282 (2018)

\bibitem{wang2023artificial}
Wang, J., Wang, S., Zhang, Y.: Artificial intelligence for visually impaired. Displays  \textbf{77},  102391 (2023)

\bibitem{WHO}
WHO: Health report on diabetic patient (2024), \url{https://tinyurl.com/bdzz54nm}

\bibitem{yi2019generative}
Yi, X., Walia, E., Babyn, P.: Generative adversarial network in medical imaging: A review. Medical image analysis  \textbf{58},  101552 (2019)

\bibitem{zhang2021classification}
Zhang, J., Xie, B., Wu, X., Ram, R., Liang, D.: Classification of diabetic retinopathy severity in fundus images with densenet121 and resnet50. arXiv preprint arXiv:2108.08473  (2021)

\end{thebibliography}
\balance
\end{document}